\documentclass[conference]{IEEEtran}
\IEEEoverridecommandlockouts
\usepackage{cite}
\usepackage{amsmath,amssymb,amsfonts}
\usepackage{gensymb}
\usepackage{algorithm}
\usepackage{algorithmic}
\usepackage{graphicx}
\usepackage{textcomp}
\usepackage{xcolor}
\usepackage[acronym]{glossaries}
\usepackage{subcaption}
\usepackage{url}
\usepackage{multirow}
\usepackage[top=1in, bottom=0.76in, left=0.75in, right=0.75in]{geometry}
\usepackage{comment}
\usepackage{pgfplots}
\usetikzlibrary{patterns}
\usepgfplotslibrary{groupplots} % LATEX and plain TEX
\usetikzlibrary{pgfplots.groupplots} % LATEX and plain TEX
\pgfplotsset{width=10cm, compat=1.9}

\tikzset{
    png export/.style={
        external/system call/.add={}{; convert -density 300 -transparent white "\image.pdf" "\image.png"},
        /pgf/images/external info,
        /pgf/images/include external/.code={%
            \includegraphics
            [width=\pgfexternalwidth,height=\pgfexternalheight]
            {##1.png}%
        },
    },
    png export,% ACTIVATE
}

\makeatletter
    \newcommand{\linebreakand}{%
      \end{@IEEEauthorhalign}
      \hfill\mbox{}\par
      \mbox{}\hfill\begin{@IEEEauthorhalign}
    }
    \makeatother

\def\BibTeX{{\rm B\kern-.05em{\sc i\kern-.025em b}\kern-.08em
    T\kern-.1667em\lower.7ex\hbox{E}\kern-.125emX}}
\begin{document}

\pgfmathdeclarefunction{gauss}{2}{%
  \pgfmathparse{1/(#2*sqrt(2*pi))*exp(-((x-#1)^2)/(2*#2^2))}%
}

\newacronym{IST}{IST}{Instituto Superior T\'ecnico}
\newacronym{GP}{GP}{Gaussian Process}
\newacronym{EKF}{EKF}{Extended Kalman Filter}
\newacronym{KF}{KF}{Kalman Filter}
\newacronym{RLS}{RLS}{Recursive Least Squares}
\newacronym{YARP}{YARP}{Yet Another Robot Platform}
\newacronym{dh}{DH}{Denavit-Hartenberg}
\newacronym{al}{AL}{Active Learning}
\newacronym{csal}{CSAL}{Cost-Sensitive Active Learning}
\newacronym{ucsal}{UCSAL}{Unconstrained Cost-Sensitive Active Learning}
\newacronym{ccsal}{CCSAL}{Constrained Cost-Sensitive Active Learning}
\newacronym{r}{R}{Random}
\newacronym{cn}{CN}{Constant Noise}
\newacronym{pdn}{PDN}{Pose Dependent Noise}

%\title{Active Robot Learning for Efficient Body-Schema Online Adaptation}

\title{Online Body Schema Adaptation through Cost-Sensitive Active Learning}

\author{Gonçalo Cunha$^{1}$, Pedro Vicente$^{1}$, Alexandre Bernardino$^{1}$, Ricardo Ribeiro$^{1}$, Plinio Moreno$^{1}$  % <-this % stops a space

\thanks{This work was partially supported by FCT with the LARSyS - FCT Project UIDB/50009/2020 and the PhD grant PD/BD/135115/2017.}
\thanks{$^{1}$ Authors are with the Institute for Systems and Robotics, Instituto Superior Técnico, Universidade de Lisboa, Lisbon, Portugal.
        {\tt\footnotesize goncaloccunha@tecnico.ulisboa.pt 
        ,
        \{pvicente,alex,ribeiro,plinio\}@isr.tecnico.ulisboa.pt},}
        %,
        %{ribeiro,plinio}@isr.tecnico.ulisboa.pt}.}%
}

\maketitle

\begin{abstract}
%This thesis proposes a movement efficient approach for estimating the Denavit-Hartenberg (DH) parameters of 7 joints of a humanoid robot arm, in a simulation environment, using the iCub simulator. A cost-sensitive active learning approach based on the A-Optimality criterion is used to select optimal joint configurations. The chosen joint configurations are informative to the estimation and minimise the movement between samples, simultaneously, thus reducing energy consumption, along with mechanical fatigue and wear, while not compromising the calibration procedure. The hand pose is sampled with the iCub eyes (cameras), using ArUco markers. The samples are used to estimate the DH parameters using an Extended Kalman Filter. Also, a non-parametric occlusion model is proposed to avoid choosing joint configurations where the markers are not visible to the cameras, to reduce the number of failed sampling attempts. The results show cost-sensitive active learning can perform similarly to the standard active learning approach, while significantly reducing the necessary movement.

Humanoid robots have complex bodies and kinematic chains with several Degrees-of-Freedom (DoF) which are difficult to model. 
Learning the parameters of a kinematic model can be achieved by observing the position of the robot links during prospective motions and minimising the prediction errors.
This work proposes a movement efficient approach for estimating online the body-schema of a humanoid robot arm in the form of Denavit-Hartenberg (DH) parameters.
 A cost-sensitive active learning approach based on the A-Optimality criterion is used to select optimal joint configurations. The chosen joint configurations simultaneously minimise the error in the estimation of the body schema and minimise the movement between samples. This reduces energy consumption, along with mechanical fatigue and wear, while not compromising the learning accuracy. 
 The work was implemented in a simulation environment, using the 7DoF arm of the iCub robot simulator.
 The hand pose is measured with a single camera via markers placed in the palm and back of the robot's hand.  
 A non-parametric occlusion model is proposed to avoid choosing joint configurations where the markers are not visible, thus preventing worthless attempts. The results show cost-sensitive active learning has similar accuracy to the standard active learning approach, while reducing in about half the executed movement.
 
\end{abstract}


\section{Introduction}
\label{section:Introduction}

%%MOTIVATION

Robots are generally deployed to have a fixed behaviour in low uncertainty environments (e.g. factories) and they rely on their body-schema to accomplish many of their tasks. Generally, robots require expensive and time consuming calibrations performed by experts since body parts: i) may not have the exact dimensions they should and ii) they may be affected by material wear and fatigue. Even with these offline calibration procedures, the presence of abnormal conditions or disturbances, such as changes in room temperature causing materials to expand or contract, may affect their performance if they lack the ability to adapt their models online. 

Humans learn their own body-schema by using sensorimotor information in a process that starts at early infancy \cite{Meer1997}. This continuous learning is what allows humans to be able to adapt to different conditions. Producing robots possessing similar behaviours is essential for many areas of robotics where extended periods of autonomous behaviour are required, such as exploration robots in inaccessible areas, rescue robots, social robots and human-robot cooperation.  

\begin{figure}
\centering
  \includegraphics[width=0.8\linewidth]{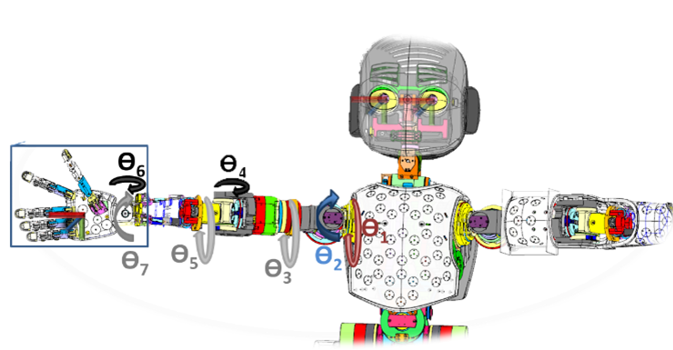}
  \caption{iCub Computer Aided Design (CAD) model.}
  \label{fig:kinematics}
\end{figure}

Online body-schema adaptation uses machine learning algorithms, which require significant amounts of data, normally acquired by performing random movements.
%and obtaining measurements. 
This may result in acquiring irrelevant data or data which does not improve the models. Active learning aims to reduce the amount of training data required to build a model, with a certain precision, by having the learning algorithm decide which data it wants to sample next. A general introduction for this area of research can be found in \cite{Settles2012}. Cost-sensitive active learning is a concept also explored in \cite{Settles2012}. The main idea is that the learning task may be associated with other costs. As an example, robots designed to perform chemical experiments, such as \cite{Burger2020AChemist}, use optimisation algorithms to decide the next experiment based on the learning potential, as well as the monetary costs. In the context of robotics and body-schema learning, requiring less data means the robot is able to adapt faster to new unpredictable conditions.

%%GOALS
The calibration problem consists of the estimation of a set of parameters associated with the arm's physical characteristics. The parameter estimation should use samples from the position and orientation of the hand, i.e. the hand pose, while knowing the readings of proprioceptive sensors (joint encoders). An active learning approach chooses the best joint configurations in order to reduce the number of samples, as well as reduce the amount of movement needed which would improve time and energy efficiency as a consequence.

The proposed methods will be compared with random selection of joint values and with a conventional active learning approach, which only aims to reduce the number of samples. The comparison will be made by assessing the accuracy of the methods on learning the body-schema, the number of samples and the amount of movement needed. 

\newgeometry{top=0.75in, bottom=0.76in, left=0.75in, right=0.75in}
\section{Related Work}
\label{section:relatedWork}
Recent works have succeeded in employing different strategies for body-schema adaptation, such as \cite{Vicente2016,Zenha2018,Stepanova2019RobotRobot}. All these works successfully adapt their body models to account for the robot's body errors, but they fail at choosing the most informative samples to do so. Using active learning could improve results and reduce the number of samples.

Active learning methods have been employed and have shown empirical success in multiple areas of robotics, such as \cite{Martinez-Hernandez2017,Martinez-Hernandez2018,Lu2020Multi-FingeredLearning,Ribes2015}. In the context of body-schema learning, \cite{Martinez-Cantin2010} used active learning to estimate a kinematic model of a serial robot and \cite{Baranes2013ActiveRobots} used an intrinsically motivated goal exploration mechanism to learn inverse models in high-dimensional redundant robots. All these works have shown the advantages of using active learning. The main focus was to minimise the number of samples, reducing the computation and learning times, however, they do not take into account the minimisation of the actual effort or movement performed by the robots. 

In \cite{Matsubara2017} and \cite{Ottenhaus2018}, the authors proposed criteria for active touch point selection to estimate object shapes which consider both error reduction and exploration costs. 
These works %support the use of 
exploited
cost-sensitive active learning, 
%since they were 
being
able to minimise the accumulated path length needed for accurate estimation with low impact on the number of touches necessary. 
This is similar to what is desired on this work
on body schema adaptation.
%since 
Indeed, the cost of an exploration action is related to the required movement to perform it. 

\subsection{Contributions}
\label{section:contributions}

This work aims to create a calibration routine which can be performed by the robot using active learning for sampling and movement efficiency. This consists of estimating the \gls{dh} parameters of the robotic arm. The estimation of these parameters will use observations of the pose (position and orientation) of the hand, using visual input. %This will be tested using the iCub arm, shown in Fig.~\ref{fig:kinematics}, which has seven rotational joints. 
This work is a follow-up on \cite{Cunha}. 
%where the observations of the hand pose were simulated geometrically, instead of using visual input. 
Instead of the previous geometrical simulated observations, we are using visual input to retrieve the hand pose.
Using visual input to obtain samples of the hand pose comes with two additional challenges. The first challenge is hand visual occlusion and the second challenge is the heteroscedastic (non-uniform) measurement error, which is due to the method used for hand pose detection in the camera images.

Therefore, the contributions of this work are threefold: i) a vision-based cost-sensitive active learning approach for body-schema learning, ii) an online hand occlusion learning method based on the previous joint configurations and iii) an observation noise predictor based on the hand pose.

In the end, we expect to select the best joint angles, $\pmb{\theta} = \begin{bmatrix} \theta^{(0)} & \theta^{(1)} & \cdots & \theta^{(n)}\end{bmatrix}$ to sample the hand pose, aiming to reduce the number of samples required and the required movement.
%Similarly to \cite{Matsubara2017} and \cite{Ottenhaus2018}, we argue that using active learning to reduce the number of samples taken may not be the best approach, since some of the best samples may require unnecessary long movements, increasing execution time and energy spent. 
%The main contribution of this work is a cost-sensitive active learning approach for body-schema learning, which chooses the best joint angles, $\pmb{\theta} = \begin{bmatrix} \theta^{(0)} & \theta^{(1)} & \cdots & \theta^{(n)}\end{bmatrix}$, to sample the hand pose, aiming to reduce the number of samples required and the required movement. The proposed calibration routine is composed of the key steps shown in Fig.~\ref{fig:programStructure}.

%Using visual input to obtain samples of the hand pose comes with two additional challenges. The first challenge is hand occlusion. To address this issue, a kernel based method is used to learn the likelihood of observing the hand at a given joint configuration, using information from the previous attempts. The second challenge is heteroscedastic (non-uniform) measurement error. This is due to the way the hand pose is detected in the camera images and image resolution limitations. Not having a rough estimate of the expected noise may lead to failed estimations. It is proposed a noise predictor for the used measurement method, based on previous studies on this matter.

\begin{figure}
\centering
  \includegraphics[width=0.9\linewidth]{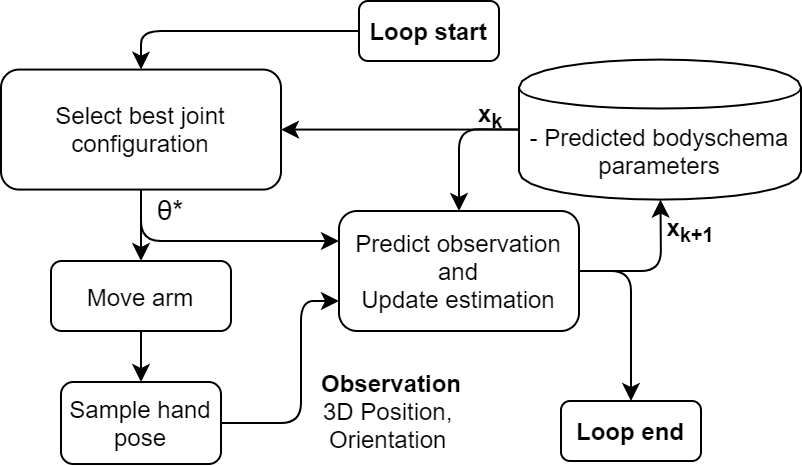}
  \caption{Key steps in the program to estimate the robot's physical parameters, $\pmb{x}$. The subscript $k$ indicates the algorithm's iteration and $\pmb{\theta}^*$ is the selected joint configuration.}
  \label{fig:programStructure}
\end{figure}

\section{Methods}
\label{section:methods}
This Section presents the main methods used to build a system capable of performing the steps in Fig.~\ref{fig:programStructure}. 
This system requires a recursive estimator to update the estimation of the \gls{dh} parameters after each sample, for which it is used an \gls{EKF}, explained in Section~\ref{section:ekf}. To guarantee sampling and movement efficiency, a cost-sensitive active learning criterion must be defined, which is presented in Section~\ref{section:activelearning}. As already mentioned, this work proposes acquiring samples of
the hand pose %for the calibration procedure
using visual input. Since the hand may not be detected by
the cameras, in Section~\ref{section:betaprocess} an occlusion model is proposed to predict whether the hand should be visible or
not. This is done with a non-parametric smoothed beta distributions learned based on previous attempts.

\subsection{Extended Kalman Filter}
\label{section:ekf}
The \gls{EKF}, explained in detail in \cite{Reid2011}, allows recursive parameter estimation of systems represented by a nonlinear model, which is the case for the relation between the \gls{dh} parameters ($\pmb{x}$) and the hand pose ($\pmb{z}$) given by the forward kinematics ($h(\pmb{x},\pmb{\theta})$) at a particular joint configuration, $\pmb{\theta}$. A measurement of the hand pose, $\pmb{z}_k$, is modelled by
\begin{equation}
    \label{eq:kalmanobs}
    \pmb{z}_k = h(\pmb{x},\pmb{\theta}_k) + \pmb{v}_k,
\end{equation}
where $\pmb{\theta}_k$ represents the joint encoder values, $\pmb{v}_k$ represents the measurement noise and $k$ is the time-step. Since $\pmb{x}$ represents the \gls{dh} parameters of the iCub's arm, it is approximately constant in time, only affected by the process noise $\pmb{w}_k$:
\begin{equation}
    \label{eq:kalmanstate2}
    \pmb{x}_{k+1} = \pmb{x}_k + \pmb{w}_k.
\end{equation}
$\pmb{v}_k$ and $\pmb{w}_k$ are zero-mean Gaussian noises.

The \gls{EKF} is divided in two steps: 1) Predict and 2) Update.

\subsubsection{Predict}
This step predicts the state $\pmb{x}$ and the prediction variance $\pmb{P}_{k+1|k}$ at time $k+1$, using only information available at time $k$. These predictions are given by:
\begin{equation}
    \label{eq:KalmanPredictState}
    \pmb{\hat{x}}_{k+1|k} = \pmb{\hat{x}}_{k|k},
\end{equation}
\begin{equation}
    \label{eq:KalmanPredictVariance}
    \pmb{\hat{P}}_{k+1|k}=\pmb{\hat{P}}_{k|k} + \pmb{Q}_k,
\end{equation}
where $\pmb{Q_k}=E[\pmb{w}_k\pmb{w}_k^T]$ is the process noise co-variance matrix.
\subsubsection{Update}
This step updates the state, $\pmb{x}$, and co-variance, $\pmb{P}$, using a combination of the prediction and the new observation obtained of the hand pose, $\pmb{z}_{k}$. They are given by:
\begin{equation}
    \label{eq:EKalmanUpdateState}
    \pmb{\hat{x}}_{k+1|k+1} = \pmb{\hat{x}}_{k+1|k} + \pmb{K}_{k+1}[\pmb{z}_{k}-h(\pmb{\hat{x}}_{k+1|k}, \pmb{\theta}_k)],
\end{equation}
\begin{equation}
    \label{eq:EKalmanUpdateVariance}
    \pmb{\hat{P}}_{k+1|k+1}=\pmb{\hat{P}}_{k+1|k}-\pmb{K}_{k+1}\pmb{S}_{k+1}\pmb{K}_{k+1}^T,
\end{equation}
where $\pmb{R}_k=E[\pmb{v}_k\pmb{v}_k^T]$ is the measurement noise co-variance, $\pmb{K}_{k+1} = \pmb{\hat{P}}_{k+1|k}\pmb{H}_{k+1}^T\pmb{S}_{k+1}^{-1}$ is the Kalman gain with $\pmb{S}_{k+1}=\pmb{H}_k\pmb{\hat{P}}_{k+1|k}\pmb{H}_{k}^T+\pmb{R}_{k+1}$, and $\pmb{H}$ is the jacobian matrix of the observation function in \eqref{eq:kalmanobs}, with respect to $\pmb{x}$, $\pmb{H} = \frac{\partial h}{\partial \pmb{x}}$.
%\begin{equation}
 %   \label{eq:EkalmanGain}
%    \pmb{K}_{k+1} = \pmb{P}_{k+1|k}\pmb{H}_{k+1}^T\pmb{S}_{k+1}^{-1},
%\end{equation}

%\begin{equation}
%    \label{eq:EKFs}
%    \pmb{S}_{k+1}=\pmb{H}_k\pmb{P}_{k+1|k}\pmb{H}_{k}^T+\pmb{R}_{k+1},
%\end{equation}

%\begin{equation}
 %   \label{eq:EKFlinearization2}
  %  H = \frac{\partial h}{\partial x}.
%\end{equation}

\subsection{Cost-Sensitive Active Learning}
\label{section:activelearning}

The goal of an active learner is to identify what is the optimal action to perform. After choosing the appropriate criterion, one must choose an appropriate cost function, $C(\cdot)$, and the next optimal action ($\pmb{\theta^*}$) is given by:
\begin{equation}
    \label{eq:ALgeneralCost}
    \begin{aligned}
        \pmb{\theta}^* = & \underset{\pmb{\theta}}{ \text{ argmin}}
        & & C(\pmb{\theta}). \\
    \end{aligned}
\end{equation}

The A-optimality criterion was proposed by \cite{Martinez-Cantin2010} for active selection of the joint angles, $\pmb{\theta}$. It consists of defining the cost function as the expected mean squared error of the robot parameters, $\pmb{x}$. Since the observation noise from \eqref{eq:kalmanobs} is considered to be Gaussian, the cost function approximates to the expected trace of the co-variance matrix of $\pmb{x}$, given the previous observations, $\pmb{z}_{1:k}$, for joint configurations $\pmb{\theta}_{1:k}$,
\begin{equation}
    \label{eq:costPmatrix}
    \begin{split}
        C_0(\pmb{\theta}) = \mathbb{E}\left[(\pmb{\hat{x}}_{k+1} - \pmb{x})^T(\pmb{\hat{x}}_{k+1} - \pmb{x})|\pmb{z}_{1:k}, \pmb{\theta}_{1:k}\right] \\
    \approx \mathbb{E}\left[tr(\pmb{\hat{P}}_{k+1}) | \pmb{z}_{1:k}, \pmb{\theta}_{1:k}\right].
    \end{split}
\end{equation}

 As seen in Fig.~\ref{fig:markersInHand}, positive values of $\pmb{x}$ for the hand position are not desirable due to the robot $\pmb{x}$ axis pointing backwards. A term was added to the cost function from \eqref{eq:costPmatrix} to penalise positive values for $\pmb{x}$ in the predicted position, $\pmb{\hat{p}}$, for a given joint configuration $\pmb{\theta}$. Therefore, it was adapted to
\begin{equation}
    \label{eq:changedActiveLearning}
    C\left(\pmb{\theta} \right) = \frac{C_0\left(\pmb{\theta} \right)}{\Bar{\eta}\left(\pmb{\theta} \right)} + a\cdot \text{arctan}\left(b\cdot \pmb{\hat{p}}_x \left(\pmb{\theta} \right)\right),
\end{equation}
where $a,b>0$ are tunable, and $\Bar{\eta}$ is the likelihood of the hand being detected by the cameras for the given joint configuration $\pmb{\theta}$. This is further explained in Section~\ref{section:betaprocess}.

In this work, the sample acquisition cost is related to the movement performed by the arm, since \eqref{eq:ALgeneralCost} selects the next configuration to where the arm moves to. This work mitigates the amount of movement performed in two separate ways.

\subsubsection{Unconstrained Optimisation}

To penalise the amount of movement in the calibration routine, one can change \eqref{eq:ALgeneralCost} to
\begin{equation}
    \label{eq:csALgeneralCost}
    \begin{aligned}
        \pmb{\theta}^*_k = & \underset{\pmb{\theta}}{\text{ argmin}}
        & & C(\pmb{\theta}) + \gamma d(\pmb{\theta}, \pmb{\theta}^*_{k-1}),\\
    \end{aligned}
\end{equation}
where $d(\pmb{\theta}, \pmb{\theta}^*_{k-1})$ is a distance, representing the cost associated to moving to the next joint configuration, $\pmb{\theta}$, accounting for the previous one, $\pmb{\theta}^*_{k-1}$, and $\gamma$ is a parameter which can be changed according to the relative weight intended.

\subsubsection{Constrained Optimisation}

Reducing the amount of arm movement during the calibration routine can also be done by constraining the optimisation problem. The choice of the next joint configuration is given by
\begin{equation}
\label{eq:ALConstrained}
   \begin{aligned}
          \pmb{\theta}^*_k =  & \underset{\pmb{\theta} \in \left[ \pmb{\theta}^*_{k-1} - \pmb{\Delta}, \pmb{\theta}^*_{k-1} + \pmb{\Delta} \right]}{\text{ argmin}}
        & & C(\pmb{\theta}), \\
    \end{aligned}
\end{equation}
where $\pmb{\theta}^*_{k-1}$ is the previous joint configuration and $\pmb{\Delta}$ is a n-dimensional vector (equal to the number of joints). Considering normalised joint values in the interval $\left[0, 1\right]$, $\pmb{\Delta}$ is defined as $\pmb{\Delta} = \delta\cdot \pmb{1_n}$, where $\pmb{1_n}$ is a unit vector of size $n$ and $\delta$ is a parameter that defines the relative movement every joint can perform around the current configuration.

\subsection{Non-parametric Occlusion Model}
\label{section:betaprocess}
When a new joint configuration is selected, the hand may not be visible to the cameras. To avoid repeated selection of joint configurations where this happens, we propose using a kernel based non-parametric occlusion model. Note, that this prediction cannot be done based on the kinematics since the errors could leave to erroneous predictions.

In \cite{Montesano2009LearningDescriptors}, a kernel based non parametric approach is used to predict the probability of a successful grasp by extrapolating past data to new unseen features. The same method is applied in this work to predict the likelihood, $\eta_*$, of having a visible marker to the cameras in a particular joint configuration, $\pmb{\theta}_*$. This is done given the past observations, $\pmb{Y}=\{y_1, y_2, \dots, y_K\}$, on the joint configurations, $\pmb{\Theta} = \{\pmb{\theta}_0, \pmb{\theta}_1, \dots, \pmb{\theta}_K\}$, where each $y_k$ contains the number of successful and unsuccessful sampling attempts, $S_k$ and $U_k$, respectively.

The Bayes rule is used in \cite{Montesano2009LearningDescriptors} to obtain the posterior
\begin{equation}
    \label{eq:posteriorSmoothBeta}
    \begin{split}
        p(\eta_*|\pmb{\theta}_*, \pmb{\Theta}, \pmb{Y}) \propto p(\pmb{Y}|\eta_*, \pmb{\theta}_*, \pmb{\Theta})p(\eta_*|\pmb{\theta}_*,\pmb{\Theta}),
    \end{split}
\end{equation}
where the prior $p(\eta_*|\pmb{\theta}_*,\pmb{\Theta})$ is modelled as a Beta distribution, with parameters $a_0$ and $b_0$, $Be(a_0,b_0)$. Based on the properties of Beta and Binomial distributions, the posterior is deduced as:
\begin{equation}
    %\begin{multline}
    \label{eq:posteriorBeta}
    \resizebox{0.89\linewidth}{!}{%
        $p(\eta_*|\pmb{\theta}_*, \pmb{\Theta}, \pmb{Y}) = \text{Be}\left(\eta_*; \sum_{k=0}^M S_{*k} + a_0, \sum_{k=0}^K U_{*k} + b_0\right),
        $
    }
    %\end{multline}
\end{equation}
where $S_{*k} = \mathcal{K}(\pmb{\theta}_*,\pmb{\theta}_k) \cdot S_k$ and $U_{*k} = \mathcal{K}(\pmb{\theta}_*,\pmb{\theta}_k) \cdot U_k$ are the predicted number of successful and failed samples at the joint configuration $\pmb{\theta}_*$, given the previous attempts at $\pmb{\theta}_k$, propagated by a kernel function, $\mathcal{K}$. 
It was used a squared exponential kernel so that the diffusion process keeps $S_* = S_k$ and $U_* = U_k$ for $\pmb{\theta}_* = \pmb{\theta}_k$ and decreases as $\pmb{\theta}_*$ is farther in the joint space. Finally, the predicted sampling success mean probability is:
\begin{equation}
\label{eq:finalbeta}
    \Bar{\eta}_* = \frac{\sum_{k=0}^K S_{*k} + a_0}{\sum_{k=0}^K S_{*k} + a_0 + \sum_{k=0}^K U_{*k} + b_0}.
\end{equation}
This result is used to select joint configurations more likely to have a visible marker in the cost function from \eqref{eq:changedActiveLearning}.

\section{Experimental Setting}
\label{section:implementation}

This Section describes the implementation and performed experiments. 
The vision based hand pose sampling method is shown in Section~\ref{section:aruco}.  
%These are placed on the back and palm of the hand, providing a method to detect its position and orientation.
Section~\ref{section:icubsim} gives some details regarding the iCub simulator and Section~\ref{section:metrics} establishes the metrics used for comparisons and results.

\subsection{ArUco Module}
\label{section:aruco}
The OpenCV ArUco module is based on the ArUco library \cite{Garrido-Jurado2014AutomaticOcclusion}, which serves for detection of square fiducial markers. The ArUco module
%\footnote{A tutorial of how to use the module is at \url{https://docs.opencv.org/trunk/d5/dae/tutorial_aruco_detection.html}.} 
is responsible for acquiring observations of the hand pose, by placing one marker on the palm of the hand and one on the back of the hand.

\subsubsection{Measurement Noise Model}
\label{section:measNoise}

\begin{figure}
\begin{subfigure}[t]{.40\linewidth}
    \centering
    \includegraphics[width=0.99\linewidth]{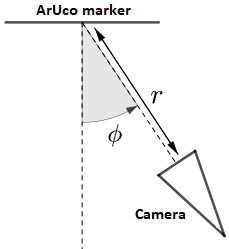}
    \caption{}
    \label{fig:cameraAngle}
\end{subfigure} \hfill%
\begin{subfigure}[t]{.55\linewidth}
    \centering
    \includegraphics[width=0.9\linewidth]{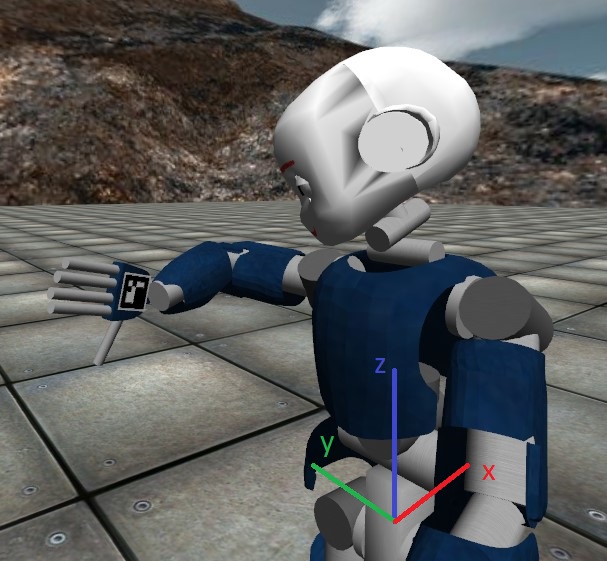}
    \caption{}
    \label{fig:markersInHand}
\end{subfigure}
\caption{ArUco marker detection by the camera in the iCub simulator. The distance from the camera to the marker is represented by $r$ and $\phi$ represents the angle between the perpendicular direction of the marker and the direction defined by the centre point of the marker and the camera position: (a) ArUco marker pose in relation to the camera; (b) ArUco marker placed on the right hand of the iCub simulator.}
\label{fig:ArUcomarkerdetection}
\end{figure}

Measurement error and variance in fiducial markers have been studied in various works, such as \cite{Abawi2004AccuracyARToolKit,Pentenrieder2006AnalysisTracking}. They show how measurement error and variance change with the distance and angle to the camera. The works achieve similar conclusions: i) The closer to the camera the fiducial marker is, the better is the pose estimation; ii) The rotation estimation seems to be worse when the marker is directly facing the camera and the systematic error and variance are lower when it is angled 30º to 50º from the camera. 
We model the measurement noise co-variance matrix as $\pmb{R}_k=\sigma_k^2 \pmb{I}$, where $\sigma_k$ represents the standard deviation of the $k$-th measurement and $\pmb{I}$ is the identity matrix. Given the studies in the mentioned works, a very rough approximation of $\sigma_k^2$ can be made as a function of camera distance, $r$, and camera angle, $\phi$, to the marker, as in Figure~\ref{fig:cameraAngle}, given by
\begin{equation}
\label{eq:measurementNoiseChange}
    \sigma_k^2 = ar^2 + b\left(\phi-45\right)^2,
\end{equation}
where $a$ and $b$ are adjustable and were selected by trial and error, with an initial guess based on the mentioned studies.

The estimated co-variance matrix $\pmb{R}_k$ will be used on the \gls{EKF} update step and to predict the joint configuration in the active learning step. 

\subsection{iCub Simulator}
\label{section:icubsim}
In order to test our approach implementation, the open-source simulator for the humanoid robot iCub will be used \cite{Tikhanoff2008}. This simulator was designed to replicate the physics and dynamics of the real robot, as close as possible. 

The iCub library\footnote{available at \url{https://github.com/robotology/icub-main}.}
provides interfaces to interact with the simulator, either by sending commands or reading information. It allows to send commands to move all joints connecting rigid bodies, contains proprioceptive information about joint encoders and possesses 2 cameras to act as eyes.

\subsection{Comparison Metrics}
\label{section:metrics}
In this Section, a few metrics will be defined for performance evaluation of the different methods.

At each iteration, a set of $N=1000$ uniformly distributed joint configurations is generated to compute the arithmetic means of the position and orientation errors.

\subsubsection{Average Position Error}
For each of the $N$ configurations, the position error is computed using the euclidean norm and the expression for the average position error results in
\begin{equation}
    \label{eq:avPosError}
\frac{1}{N}\sum_{i=1}^N\Vert \pmb{p}_i-\pmb{\hat{p}}_i \Vert,
\end{equation}
where $p_i$ and $\hat{p}_i$ are the actual and predicted 3D hand positions, respectively.

\subsubsection{Average Orientation Error}
For each of the $N$ configurations, the orientation error is computed using:
\begin{equation}
    \label{eq:rotationError}
    d(\pmb{O}_i, \pmb{\hat{O}}_i) = \sqrt{\frac{\Vert logm(\pmb{O}_i^T\pmb{\hat{O}}_i)\Vert^2_F}{2}}\cdot\frac{180}{\pi} \quad \text{[º]},
\end{equation}
where $logm$ is the principal matrix logarithm and $\Vert \cdot \Vert_F$ is the Frobenius norm. $\pmb{O}_i$ and $\pmb{\hat{O}}_i$ are the actual and predicted hand rotation matrices. The average error results in 
\begin{equation}
    \label{eq:avRotError}
    \frac{1}{N}\sum_{i=1}^N d(\pmb{O}_i, \pmb{\hat{O}}_i) \quad \text{[º]}.
\end{equation}

\subsubsection{Accumulated Joint Movement}
At iteration $K$, the total arm movement is given by the sum of all past joint movements,
\begin{equation}
    \label{eq:totalArmMovement}
    \sum_{k=1}^K \left\Vert \pmb{\theta}_k - \pmb{\theta}_{k-1}\right\Vert_1 \quad \text{[º]},
\end{equation}
where $\Vert\cdot\Vert_1$ represents the $l1$ norm and $\pmb{\theta}_0$ is the initial joint configuration.

\section{Results}
\label{section:results}

The calibration routine from Fig.~\ref{fig:programStructure} was performed using four methods for selecting the best joint configuration.
%the joint value selection of the arm.
%, which are summarised in Table~\ref{tab:methodSummary}. 
The first is \gls{r}, which selects random joint configurations uniformly distributed, which usually serves as a base line for comparison with active learning. The second method is the \gls{al} method, which does not consider movement costs
(solving \eqref{eq:ALgeneralCost}).
The third and fourth methods are the cost-sensitive approaches proposed in this work, \gls{ucsal} (solving \eqref{eq:csALgeneralCost}) and \gls{ccsal} (solving \eqref{eq:ALConstrained}). Solving the optimisation problems 
%\eqref{eq:ALgeneralCost}, \eqref{eq:csALgeneralCost} and \eqref{eq:ALConstrained}, 
requires a global optimisation algorithm, since the cost functions are not easily differentiable. It is used the NLopt \cite{Johnson2011} implementation of the DIRECT algorithm \cite{Jones1993LipschitzianConstant}.

% \begin{table}
% \centering
% \begin{tabular}{cc}
% \textbf{Method} & \textbf{Joint Selection} \\ \hline
% Random (\textbf{R}) & Uniform random selection \\ \hline
% Active Learning (\textbf{AL}) & Solves \eqref{eq:ALgeneralCost} \\ \hline
% \begin{tabular}{c}
%     Unconstrained Cost-Sensitive   \\
%     Active Learning (\textbf{UCSAL})  
% \end{tabular}  & Solves \eqref{eq:csALgeneralCost} \\ \hline
% \begin{tabular}{c}
%     Constrained Cost-Sensitive   \\
%     Active Learning (\textbf{CCSAL})  
% \end{tabular} & Solves \eqref{eq:ALConstrained}
% \end{tabular}%
% \caption{Summary of the different used joint selection methods.}
% \label{tab:methodSummary}
% \end{table}

A random initial joint configuration for the right arm of the robot is set and the algorithm from Fig.~\ref{fig:programStructure} executes all four different methods. 
The main loop of Fig.~\ref{fig:programStructure} runs for 50 iterations selecting a new sample of the hand pose based on the best configuration criteria. To obtain solid conclusions, the results displayed are the average of 50 repetitions of each experiment. At each experiment, the \gls{dh} parameters of the robot are initialised with values from a uniform distribution, where the means are the actual values of the \gls{dh} parameters and the width of the distributions are 46~mm and 54\degree~for the linear and angular parameters, respectively.

\subsection{Results Analysis}
\label{section:markerResults}

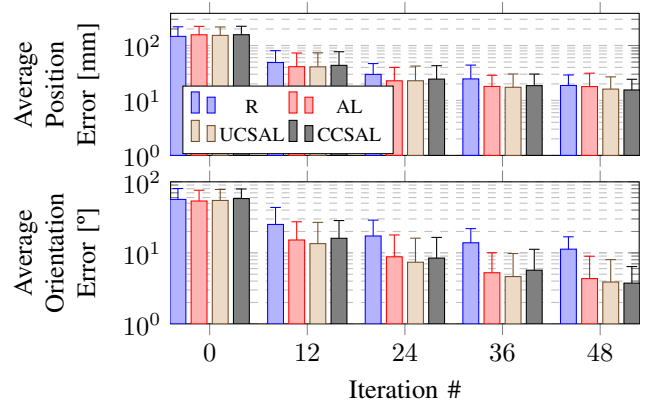
\begin{figure}
\centering
\begin{tikzpicture}

\begin{groupplot}[group style={
group size=1 by 2,
xlabels at=edge bottom,
xticklabels at=edge bottom,
ylabels at=edge left,
vertical sep=10pt,
},
xlabel={Iteration \#},
    ylabel={\begin{tabular}{c}Average\\Position \\ Error [mm]\end{tabular}},
width=0.9\linewidth,
   height=0.4\linewidth,
   ymajorgrids=true,
   yminorgrids=true,
   grid style=dashed,
   legend style={at={(0.025,0.25)},anchor=west, nodes={scale=0.75, transform shape}},
   legend columns=2,
   ymin=1, 
   ymode=log,
   xtick={0,12,24,36,48}
 ]   
\nextgroupplot[ybar, bar width=6pt]

%%%BLUE

\addplot+[
    each nth point={12},error bars/.cd, y dir=plus, y explicit
    ]
    table [x expr=\coordindex, y=mean, y error=stddev]{positerationsMarkers20random.txt};

%%%ORANGE
   
\addplot+[
    each nth point={12},error bars/.cd, y dir=plus, y explicit
    ]
    table [x expr=\coordindex, y=mean, y error=stddev]{positerationsMarkers200DIRECTl100.000000.txt};

%%%Green
\addplot+[
    each nth point={12},error bars/.cd, y dir=plus, y explicit
    ]
    table [x expr=\coordindex, y=mean, y error=stddev]{positerationsMarkers200DIRECTl100.000010.txt};
    
%%%RED
  
\addplot+[
    each nth point={12},error bars/.cd, y dir=plus, y explicit
    ]
    table [x expr=\coordindex, y=mean, y error=stddev]{positerationsMarkers200DIRECTl100.500000.txt};

\legend{R, AL, UCSAL, CCSAL}

\nextgroupplot[ylabel={\begin{tabular}{c}Average\\Orientation \\ Error [º]\end{tabular}}, ymax=100, ybar, bar width=6pt]
%%%BLUE

\addplot+[
    each nth point={12},error bars/.cd, y dir=plus, y explicit
    ]
    table [x expr=\coordindex, y expr=\thisrow{mean}*180/3.14159265359, y error expr=\thisrow{stddev}*180/3.14159265359]{rotiterationsMarkers20random.txt};

%%%ORANGE
   
\addplot+[
    each nth point={12},error bars/.cd, y dir=plus, y explicit
    ]
    table [x expr=\coordindex, y expr=\thisrow{mean}*180/3.14159265359, y error expr=\thisrow{stddev}*180/3.14159265359]{rotiterationsMarkers200DIRECTl100.000000.txt};

%%%green
  
\addplot+[
    each nth point={12},error bars/.cd, y dir=plus, y explicit
    ]
    table [x expr=\coordindex, y expr=\thisrow{mean}*180/3.14159265359, y error expr=\thisrow{stddev}*180/3.14159265359]{rotiterationsMarkers200DIRECTl100.000010.txt};
    
%%%RED
  
\addplot+[
    each nth point={12},error bars/.cd, y dir=plus, y explicit
    ]
    table [x expr=\coordindex, y expr=\thisrow{mean}*180/3.14159265359, y error expr=\thisrow{stddev}*180/3.14159265359]{rotiterationsMarkers200DIRECTl100.500000.txt};

\end{groupplot}

\end{tikzpicture}

\caption{Results with respect to the loop iterations of the mean position and orientation errors over 50 experiments using the different joint selection methods. Baselines: R-Random, AL-Conventional AL; Ours: CCSAL-Constrained Cost-Sensitive AL and UCSAL-Unconstrained Cost-Sensitive AL.}
\label{fig:markersResultsErrorIterations}

\end{figure}

By observing the position error on Fig.~\ref{fig:markersResultsErrorIterations}, the \gls{r} method and the others perform similarly. This may be due to the limitations imposed by the pose measurement method. The fiducial markers have a mean error of about 1~cm and there is not much of a margin to improve it further. This could be improved by using cameras with a better resolution, which are not available in the iCub. Oppositely, the orientation errors on Fig.~\ref{fig:markersResultsErrorIterations} show the improved performance of \gls{al}, \gls{ucsal} and \gls{ccsal}. The \gls{r} method is worse at estimating the hand orientation than the active learning methods, since the samples are less informative, resulting in small error decrements.
%
%Fig.~\ref{fig:markersResultsErrorIterations} shows the methods \gls{al}, \gls{ucsal} and \gls{ccsal} perform very similarly. This means the cost-sensitive methods are providing samples just as informative as the \gls{al} method. In the results from Fig.~\ref{fig:markersResultsErrorTravel}, \gls{ucsal} and \gls{ccsal} reveal to be much more advantageous since it took, roughly, half the movement and the same amount of iterations to achieve similar results. When comparing the two cost-sensitive methods, their lines almost overlap in the error plots from Figs.~\ref{fig:markersResultsErrorIterations}~and~\ref{fig:markersResultsErrorTravel}, indicating that both methods allow similar amounts of exploration and exploitation. 
%
Fig.~\ref{fig:markersResultsErrorIterations} shows the methods \gls{al}, \gls{ucsal} and \gls{ccsal} perform similarly. This means samples from the cost-sensitive methods are as informative as the conventional \gls{al}. 

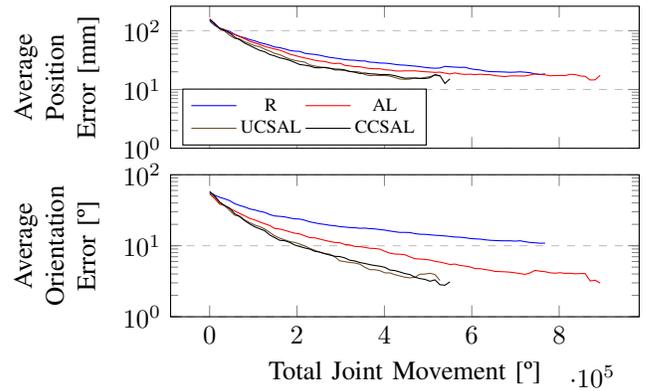
\begin{figure}
\centering
\begin{tikzpicture}

\begin{groupplot}[group style={
group size=1 by 2,
xlabels at=edge bottom,
xticklabels at=edge bottom,
ylabels at=edge left,
vertical sep=10pt,
},
xlabel={Total Joint Movement [º]},
    ylabel={\begin{tabular}{c}Average\\Position\\ Error [mm]\end{tabular} },
width=0.9\linewidth,
height=0.4\linewidth,
   ymajorgrids=true,
   grid style=dashed,
   ymin=1,
   legend style={at={(0.025,0.22)},anchor=west,
   legend columns=2,
   nodes={scale=0.7, transform shape}}
 ]   
\nextgroupplot[ymode=log]

%%%BLUE

\addplot+[
    mark=dot,
    ]
    table [x expr=200*\coordindex*180/3.14159265359 , y=mean]{postravelMarkers20random.txt};

%%%ORANGE
   
\addplot+[
    mark=dot,
    ]
    table [x expr=200*\coordindex*180/3.14159265359 , y=mean]{postravelMarkers200DIRECTl100.000000.txt};

%%%green
  
\addplot+[
    mark=dot,
    ]
    table [x expr=200*\coordindex*180/3.14159265359 , y=mean]{postravelMarkers200DIRECTl100.000010.txt};
    
%%%RED
  
\addplot+[
    mark=dot,
    ]
    table [x expr=200*\coordindex*180/3.14159265359 , y=mean]{postravelMarkers200DIRECTl100.500000.txt};

\legend{R, AL, UCSAL, CCSAL}

\nextgroupplot[ylabel={\begin{tabular}{c}Average\\Orientation \\ Error [º]\end{tabular}}, ymode=log, ymax=100 ]
%%%BLUE

\addplot+[
    mark=dot,
    ]
    table [x expr=200*\coordindex*180/3.14159265359 , y expr=\thisrow{mean}*180/3.14159265359]{rottravelMarkers20random.txt};

%%%ORANGE
   
\addplot+[
    mark=dot,
    ]
    table [x expr=200*\coordindex*180/3.14159265359 , y expr=\thisrow{mean}*180/3.14159265359]{rottravelMarkers200DIRECTl100.000000.txt};

%%%green
   
\addplot+[
    mark=dot,
    ]
    table [x expr=200*\coordindex*180/3.14159265359 , y expr=\thisrow{mean}*180/3.14159265359]{rottravelMarkers200DIRECTl100.000010.txt};    
    
    %%%RED
  
\addplot+[
    mark=dot,
    ]
    table [x expr=200*\coordindex*180/3.14159265359 , y expr=\thisrow{mean}*180/3.14159265359]{rottravelMarkers200DIRECTl100.500000.txt};

\end{groupplot}

\end{tikzpicture}

\caption{Results with respect to accumulated movement of the mean position and orientation errors over 50 experiments using different joint selection methods. Baselines: R-Random, AL-Conventional AL; Ours: CCSAL-Constrained Cost-Sensitive AL and UCSAL-Unconstrained Cost-Sensitive AL.}
\label{fig:markersResultsErrorTravel}

\end{figure}

In Fig.~\ref{fig:markersResultsErrorTravel}, \gls{ucsal} and \gls{ccsal} reveal to be more efficient since the arm movement is, roughly, half of the movement of \gls{al} without compromising the performance. 
Regarding the two cost-sensitive methods, their lines almost overlap in the plots from Figs.~\ref{fig:markersResultsErrorIterations}~and~\ref{fig:markersResultsErrorTravel}, indicating that both methods allow similar amounts of exploration and exploitation.

%%%%%%%ITERATIONS

%%%%%TRAVEL

\subsubsection{Predicting Measurement Noise}
\label{section:resultsPredictingNoise}

This Section shows the impact of the proposed noise model from \eqref{eq:measurementNoiseChange} on the parameter estimation. 
%
% \begin{table}
% \centering
% \begin{tabular}{lcc}
% \hline
% \multicolumn{1}{l}{\multirow{2}{*}{\textbf{Method}}} & \multicolumn{2}{c}{\textbf{Measurement Noise Model}} \\ \cline{2-3} 
% \multicolumn{1}{c}{} & \textbf{EKF} & \textbf{Joint Value Selection} \\ \hline
% \textbf{Naive EKF} & No & No \\ \hline
% \textbf{Naive AL} & Yes & No \\ \hline
% \textbf{AL} & Yes & Yes \\ \hline
% \end{tabular}
% \caption{Summary of the different experiments performed to evaluate the impact of the measurement noise model.}
% \label{tab:noiseModelExperiments}
% \end{table}
%
%
%Without losing generality, we will be using the standard AL approach, solving \eqref{eq:ALgeneralCost}, since it is the core cost for all the methods.
As explained in Section~\ref{section:measNoise} our system uses a pose dependent noise predictor. In this Section, it will be compared to a system where it is assumed the measurement noise is constant. This separate implementation is referred to as \gls{cn}. It will be tested for all joint selection methods, \gls{al}-\gls{cn}, \gls{ucsal}-\gls{cn},  \gls{ccsal}-\gls{cn}, and compared with our implementations using the pose dependent noise model, \gls{al}, \gls{ucsal} and \gls{ccsal}.

%All three methods use active learning, but they are not cost-sensitive. The results still apply to the general case, since all methods are using the same active learning approach of solving \eqref{eq:ALgeneralCost}. 

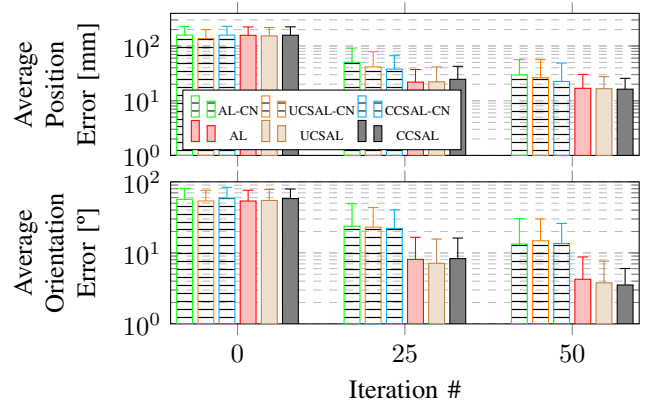
\begin{figure}
\centering
\begin{tikzpicture}

\begin{groupplot}[group style={
group size=1 by 2,
xlabels at=edge bottom,
xticklabels at=edge bottom,
ylabels at=edge left,
vertical sep=10pt,
},
xlabel={Iteration \#},
    ylabel={\begin{tabular}{c}Average\\Position \\ Error [mm]\end{tabular}},
    width=0.9\linewidth,
    height=0.4\linewidth,
   ymajorgrids=true,
   yminorgrids=true,
   grid style=dashed,
   legend style={at={(0.025,0.25)},anchor=west, nodes={scale=0.5, transform shape}},
   legend columns=3,
   ymin=1, 
   ymode=log,
   xtick={0,25,50},
   xmin=-10,
   xmax=60
 ]   
\nextgroupplot[ybar, bar width=6pt]
%%%AL-CN
\addplot+[
    color=green, fill=green!25,pattern=horizontal lines,each nth point={25},error bars/.cd, y dir=plus, y explicit
    ]
    table [x expr=\coordindex, y=mean, y error=stddev]{positerationsMarkers200NaiveDIRECTl100.000000.txt};

%%%UCSAL-CN
   
\addplot+[
    color=orange,fill=orange!25,pattern=horizontal lines,each nth point={25},error bars/.cd, y dir=plus, y explicit
    ]
    table [x expr=\coordindex, y=mean, y error=stddev]{positerationsMarkersNaiveEKFDIRECTl100.000010.txt};

%%%CCSAL-CN

\addplot+[
    color=cyan,fill=cyan!25,pattern=horizontal lines,each nth point={25},error bars/.cd, y dir=plus, y explicit
    ]
    table [x expr=\coordindex, y=mean, y error=stddev]{positerationsMarkersNaiveEKFDIRECTl100.500000.txt};

%%%%AL
   
\addplot+[
    color=red,fill=red!25,each nth point={25},error bars/.cd, y dir=plus, y explicit
    ]
    table [x expr=\coordindex, y=mean, y error=stddev]{positerationsMarkers200DIRECTl100.000000.txt};
    
%%%UCSAL
\addplot+[
    color=brown,fill=brown!25,each nth point={25},error bars/.cd, y dir=plus, y explicit
    ]
    table [x expr=\coordindex, y=mean, y error=stddev]{positerationsMarkers200DIRECTl100.000010.txt};
    
%%%CCSAL
  
\addplot+[
    color=black,fill=black!50,each nth point={25},error bars/.cd, y dir=plus, y explicit
    ]
    table [x expr=\coordindex, y=mean, y error=stddev]{positerationsMarkers200DIRECTl100.500000.txt};

    \legend{AL-CN, UCSAL-CN, CCSAL-CN, AL, UCSAL, CCSAL}

\nextgroupplot[ylabel={\begin{tabular}{c}Average\\Orientation \\ Error [º]\end{tabular}}, ymax=100, ybar, bar width=6pt]

%%%AL-CN

\addplot+[
    color=green,fill=green!25,pattern=horizontal lines,each nth point={25},error bars/.cd, y dir=plus, y explicit
    ]
    table [x expr=\coordindex, y expr=\thisrow{mean}*180/3.14159265359, y error expr=\thisrow{stddev}*180/3.14159265359]{rotiterationsMarkers200NaiveDIRECTl100.000000.txt};

%%%UCSAL-CN
   
\addplot+[
    color=orange,fill=orange!25,pattern=horizontal lines,each nth point={25},error bars/.cd, y dir=plus, y explicit
    ]
    table [x expr=\coordindex, y expr=\thisrow{mean}*180/3.14159265359, y error expr=\thisrow{stddev}*180/3.14159265359]{rotiterationsMarkersNaiveEKFDIRECTl100.000010.txt};
    
%%CCSAL-CN
    
\addplot+[
    color=cyan,fill=cyan!25,pattern=horizontal lines,each nth point={25},error bars/.cd, y dir=plus, y explicit
    ]
    table [x expr=\coordindex, y expr=\thisrow{mean}*180/3.14159265359, y error expr=\thisrow{stddev}*180/3.14159265359]{rotiterationsMarkersNaiveEKFDIRECTl100.500000.txt};

%%%AL
   
\addplot+[
    color=red,fill=red!25,each nth point={25},error bars/.cd, y dir=plus, y explicit
    ]
    table [x expr=\coordindex, y expr=\thisrow{mean}*180/3.14159265359, y error expr=\thisrow{stddev}*180/3.14159265359]{rotiterationsMarkers200DIRECTl100.000000.txt};

%%%UCSAL
  
\addplot+[
    color=brown,fill=brown!25,each nth point={25},error bars/.cd, y dir=plus, y explicit
    ]
    table [x expr=\coordindex, y expr=\thisrow{mean}*180/3.14159265359, y error expr=\thisrow{stddev}*180/3.14159265359]{rotiterationsMarkers200DIRECTl100.000010.txt};
    
%%%CCSAL
  
\addplot+[
    color=black,fill=black!50,each nth point={25},error bars/.cd, y dir=plus, y explicit
    ]
    table [x expr=\coordindex, y expr=\thisrow{mean}*180/3.14159265359, y error expr=\thisrow{stddev}*180/3.14159265359]{rotiterationsMarkers200DIRECTl100.500000.txt};
\end{groupplot}
\end{tikzpicture}

\caption{Comparison of the proposed noise model from Section~\ref{section:measNoise} in the EKF and active joint selection against a constant noise solution (CN) for the three AL methods.}
\label{fig:resultsNoisePrediction}

\end{figure}

%In Naive EKF, the measurement noise co-variance matrix is constant through the entirety of the run, $\pmb{R} = \sigma^2\pmb{I}$ (the common approach). The Naive AL method uses the noise model from \eqref{eq:measurementNoiseChange} to change $\pmb{R}$ only for the update step of the \gls{EKF}. The AL method uses the noise model from \eqref{eq:measurementNoiseChange} to change $\pmb{R}$ for the update step of the \gls{EKF} and to predict the best sample in the active learning step by computing \eqref{eq:costPmatrix} using the predicted $\pmb{R}$ matrix given by the noise model for the expected hand pose.
%All the methods are summarised in Table~\ref{tab:noiseModelExperiments}. 
The average position and orientation errors are plotted in Fig~\ref{fig:resultsNoisePrediction}, where 
%
%In Fig.~\ref{fig:resultsNoisePrediction}, 
it is possible to see the differences in performance for both conditions. The \gls{cn} approach constantly performs worse than the corresponding methods using our pose dependent noise model, since it is not capable of reducing the impact of measurements more likely to be wrong. This shows that, even though \eqref{eq:measurementNoiseChange} is a very rough approximation, it was critical to achieve an improved performance.
%The Naive AL approach performs slightly better, but by not predicting the expected error in a sample before selecting the next joint configuration, the \gls{EKF} converges slower, since the samples obtained are more likely to be less reliable. 

\subsubsection{Marker Occlusion}
\label{section:markerocclusion}

\begin{figure}
    \centering
        \begin{tikzpicture}
        \begin{axis}[
        	ylabel={\begin{tabular}{c}  Average \# \\ of discarded \\ samples\end{tabular}},
        	ybar,
        	xtick style={draw=none},
        	xtick=\empty,
        	bar width=20pt,
        	width=0.9\linewidth,
        	height = 0.4\linewidth,
        	legend style={at={(0.75,0.5)},anchor=west},
        	ymin=0,
        	ymajorgrids=true
        ]
        \addplot+[error bars/.cd, y dir=both, y explicit]
        	table [x expr=1, y=mean, y error=stddev ]{discardedSamples20random.txt};
        \addplot+[error bars/.cd, y dir=both, y explicit] 
        	table [x expr=1, y=mean, y error=stddev]{discardedSamples200DIRECTl100.000000.txt};
        \addplot+[error bars/.cd, y dir=both, y explicit]
        	table [x expr=1, y=mean, y error=stddev]{discardedSamples200DIRECTl100.000010.txt};
        \addplot+[error bars/.cd, y dir=both, y explicit] 
        	table [x expr=1, y=mean, y error=stddev]{discardedSamples200DIRECTl100.500000.txt};
        \legend{R, AL, UCSAL, CCSAL}
        \end{axis}
        \end{tikzpicture}
    \caption{Average number of discarded samples due to marker occlusion for each joint selection method. The Occlusion Model reduces the number of missed detections.}
    \label{fig:discardedSamples}
\end{figure}
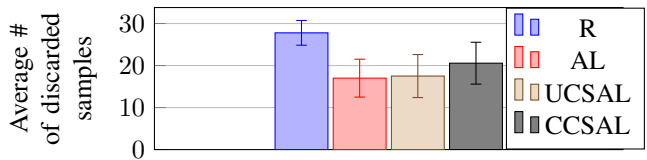

Some statistics were obtained regarding the number of discarded samples in the 50 iterations of the calibration routine and they are shown in Figure~\ref{fig:discardedSamples}. 
%These are discarded due to occlusion of the marker or due to the ArUco module not being able to locate the marker in the camera images.
These are discarded due to missed detections associated with marker occlusion or failing of the ArUco module.

Figure~\ref{fig:discardedSamples} shows the active learning methods tend to suggest fewer arm configurations where the marker is not detected. This shows the impact of the smooth beta distributions, explained in Section~\ref{section:betaprocess}. It is crucial to avoid failed sampling attempts, since, after a few failed attempts, it is able to discourage those and the surrounding joint configurations in the joint selection step. The \gls{ccsal} method seems to discard slightly more samples than the other active learning methods. Since \gls{ccsal} reduces the search bounds to reduce movement, occasionally, the search space may be reduced to one where it is harder to find the markers, while the other methods are allowed to leave those regions quickly.   

%To improve these results, the smooth beta distribution could be trained before the calibration routine, but having too much training data could increase the computation time significantly, since \eqref{eq:posteriorSmoothBeta} has complexity $\mathcal{O}(n)$ and the DIRECT algorithm must compute it for every point it evaluates.

\section{Conclusions}

This work proposed a cost-sensitive active learning approach to estimate the \gls{dh} parameters of 7 joints of the iCub arm in order to prioritise movement efficiency. Using ArUco markers provided an insight to a more realistic setting, where the samples obtained are impacted by measurement error.
The results show there is an advantage in discouraging or restricting movement during the optimisation stage. It is possible to reduce the movement performed by roughly half and still maintain the iteration wise performance. If movement efficiency is a priority, one can restrict the movement even more, at the cost of more iterations. It is worth mentioning, more iterations does not mean lower time-efficiency, since reducing the amount of time spent moving may make up for the extra computing time. Indeed, it will depend on the computing power and the speed at which the arm moves. 

For future work, it would be interesting to proceed to an implementation on the actual iCub robot to evaluate how the methods perform in the real world. It could also be built a better model to predict the observation error to check if it could improve the results significantly. After failed sampling attempts, an exploration strategy could be added to try finding the markers using head or arm movements. Redundant measurements could be used with both iCub cameras to add more robustness to noise, which could also help reducing the number of failed sampling attempts.

\bibliographystyle{IEEEtran}
\bibliography{references}

\end{document}